# Optimizing Interplanetary Trajectories with a Levy-Enhanced Hybrid Meta-heuristic Algorithm

Amin Abdollahi Dehkordi[1] and Mehdi Neshat[2]

[1]    Department of Computer Engineering, Najaf Abad Azad University, Najafabad, Isfahan, Iran

amin.abdollahi.dehkordi@gmail.com

[2]    Faculty of Engineering and Information Technology, University of Technology Sydney, Ultimo, Sydney, 2007, NSW, Australia

mehdi.neshat@uts.edu.au

Abstract. This paper proposes an advanced hybrid optimization (GMPA) algorithm to effectively address the inherent limitations of the Grey Wolf Optimizer (GWO) when applied to complex optimization scenarios. Specifically, GMPA integrates essential features from the Marine Predators Algorithm (MPA) into the GWO framework, enabling superior performance through enhanced exploration and exploitation balance. The evaluation utilizes the GTOPX benchmark dataset from the European Space Agency (ESA), encompassing highly complex interplanetary trajectory optimization problems characterized by pronounced nonlinearity and multiple conflicting objectives reflective of real-world aerospace scenarios. Central to GMPA's methodology is an elite matrix, borrowed from MPA, designed to preserve and refine high-quality solutions iteratively, thereby promoting solution diversity and minimizing premature convergence. Furthermore, GMPA incorporates a three-phase position updating mechanism combined with Lévy flights and Brownian motion to significantly bolster exploration capabilities, effectively mitigating the risk of stagnation in local optima. GMPA dynamically retains historical information on promising search areas, leveraging the memory storage features intrinsic to MPA, facilitating targeted exploitation and refinement. Empirical evaluations demonstrate GMPA's superior effectiveness compared to traditional GWO and other advanced metaheuristic algorithms, achieving markedly improved convergence rates and solution quality across GTOPX benchmarks. Consequently, GMPA emerges as a robust, efficient, and adaptive optimization approach particularly suitable for high-dimensional and complex aerospace trajectory optimization, offering significant insights and practical advancements in hybrid metaheuristic optimization techniques.

Keywords: Aerospace optimization · Interplanetary trajectory optimization · GTOPX dataset · Grey Wolf Optimizer (GWO) · Marine Predator Algorithm (MPA).

## 1    Introduction

The optimization of interplanetary trajectories constitutes a multifaceted endeavour [5]. The dynamical system exhibits nonlinearity as a consequence of the gravitational forces exerted by both the Sun and the planets. The trajectories generally encompass discontinuities in the state variables, which arise from rocket engine manoeuvres and gravitational assists throughout the trajectory [26]. The ever-evolving environment, resulting from planetary motion around the Sun, perpetuates variations in the locations of potential gravitational assists as a function of time. Additionally, the initial and/or terminal conditions undergo modifications due to this dynamic motion. A substantial array of variables is typically necessitated to accurately model trajectories that incorporate an adequate number of gravitational assists and rocket burns [19]. Reducing the number of variables generally implies that only suboptimal trajectories can be identified. The complexity of the problem is exacerbated by the fact that the fundamental configuration of gravitational assists and manoeuvres is frequently indeterminate, and the quantity of gravitational assists and manoeuvres fluctuates from one trajectory to another [24]. In order to systematically explore all conceivable trajectories, certain assumptions are conventionally employed during the preliminary phase of optimizing high-thrust interplanetary trajectories [15]. Perturbations can generally be disregarded in first-order analyses. The absolute magnitude of perturbing forces is typically significantly inferior to that of the principal forces. The Sun serves as the predominant force for the majority of the trajectory. The gravitational influence of planets only becomes significant in proximity to planetary encounters. Given that the duration of these planetary encounters is minimal, the patched-conics assumption is customarily applied. This entails modelling the trajectory within the heliocentric reference frame while representing gravitational assists as instantaneous manoeuvres within the pertinent planet-centric reference frames [26]. Likewise, due to the brevity of high-thrust impulses, these are also modelled as instantaneous changes in velocity. These first-order analyses are subsequently utilized as preliminary estimations in more comprehensive trajectory studies. Perturbations, including third-body effects, irregularities in gravitational fields, solar radiation pressure, and aerodynamic forces, must be incorporated into these more refined analyses. Moreover, additional factors may significantly influence the determination of the final trajectory, such as the potential to explore supplementary asteroids or the timing requirements associated with critical manoeuvres. In the preceding two decades, aerospace engineers have availed themselves of the opportunity to frame intricate interplanetary trajectory design challenges as global optimization problems. In contemporary practice, the aim of trajectory design encompasses not merely identifying a singular solution but also striving to ascertain the optimal solution in relation to propellant consumption whilst concurrently fulfilling the mission objectives [16]. The most pertinent Global Trajectory Optimization Problems (GTOP) pertain to Multiple Gravity Assist missions, which may incorporate the utilization of a Deep Space Maneuver (MGADSM) [2,5]. The spacecraft is capable of employing its propulsion systems to rectify its trajectory for an optimized fly-by, which justifies the expenditure of propellant. Resolutions to such realistic challenges are appropriate for conducting preliminary quantitative assessments for actual space missions; however, they necessitate the optimization of high-dimensional continuous problems characterized by a significant number of local minima. The design of optimal trajectories for interplanetary space missions constitutes a formidable research avenue within the aerospace sector and associated fields, encompassing evolutionary and metaheuristic optimization methodologies. Since 2005, the European Space Agency (ESA) has sustained the GTOP database, a compilation of numerical black-box optimization challenges aimed at the design of trajectories for real-world space missions [5]. It is noteworthy that the term "black box" pertains here to an optimization issue wherein the specific formulation of the problem remains unknown, inaccessible, or irrelevant to the optimizer. Such black-box problems frequently arise in various domains wherein intricate computer simulations are applied.

The GTOP database has garnered considerable scholarly attention, with numerous results disseminated through publication. For example, Evolutionary Algorithms (EAs) are categorized as metaheuristic strategies that have effectively addressed complex combinatorial and continuous optimization challenges in realworld problems [17]. Consequently, these algorithms have been employed in the optimization of spacecraft trajectories, with initial applications integrating Genetic Algorithms (GAs) alongside gradient-based techniques and other global optimization methods like PSO (see, e.g., [23,29]). Kim [20] and Abdelkhalik [11] have recently implemented GAs to resolve N-impulse orbital transfer challenges. The hybridization of metaheuristics has also been explored by Sentinella et al., who initially combined Differential Evolution (DE) with GAs [1], subsequently extending this approach to include Particle Swarm Optimization (PSO) [27]. In the present study, our proposed methodology is developed and assessed using an enhanced and updated iteration of the GTOP database, referred to as GTOPX, which serves as a successor to the now-defunct GTOP database. Notably, most scholarly articles addressing outcomes related to GTOP predominantly concentrate on a limited number of problem instances. This tendency arises from the inherent complexity of these challenges, which frequently necessitate millions of function evaluations for an optimization algorithm to converge upon the bestknown configurations, such as the most effective control parameters in renewable energy [9], path planning [18], etc. The intricacies associated with these benchmark problems render them both fascinating and genuinely challenging, often necessitating the deployment of massively distributed computing resources via supercomputers in the most arduous scenarios [26], [21]. While the foundational GTOP database primarily emphasized the intrinsic nature of the applications, the GTOPX collection aspires to cater to a wider audience within the community of numerical optimization researchers [28]. Metaheuristics represent sophisticated methodologies employed for the thorough investigation of optimization challenges aimed at identifying solutions that are near-optimal [25]. These methodologies are generally classified into two main categories: local searchbased algorithms and population-based algorithms, with further subdivisions encompassing evolutionary algorithms, physical algorithms, chemical algorithms, human-based algorithms, and swarm intelligence (SI) algorithms [22].

The Grey Wolf Optimizer (GWO) [14] represents a sophisticated swarm intelligence (SI) algorithm meticulously crafted to emulate the predatory behaviours exhibited by grey wolf packs in their natural habitats. It demonstrates remarkable adaptability [13], operates without the necessity of parameters [3], avoids reliance on derivatives [10], exhibits memory-less characteristics, minimizes computational demands, maintains flexibility, and upholds soundness and completeness. GWO is capable of addressing optimization challenges across a diverse array of disciplines, encompassing engineering, networking, image processing, robotics, mathematics, bioinformatics, and biomedical sciences. Considerable modifications have been implemented to effectively contend with the intricacies inherent to complex search space characteristics.

In the present investigation, GWO has been hybridized with the Marine Predators Algorithm (MPA) to mitigate its fundamental limitations, such as the propensity to become entrenched in local optima and the tendency for premature convergence; the proposed methodology is employed to resolve benchmark challenges associated with GTOPX. By leveraging the advantages of MPA, including the strategic exploration of the solution space (which involves partitioning the search area into three distinct phases to enhance both exploration and exploitation), the FDAs strategy, the retention of optimal solutions within a memory matrix, and the incorporation of a localized search strategy post-iteration through the utilization of the best solution identified thus far, followed by an update of the wolves' positions, it is anticipated that the GMPA variant will surpass the performance of the original GWO.

This manuscript is structured as follows: The subsequent section succinctly delineates the description of the GTOPX benchmark. In Section 3, we proceed to elaborate on the methodology and provide a detailed account of the GMPA, followed by an analysis of the experimental results in Section 4. Lastly, the concluding section encapsulates our findings and future research directions.

## 2  GTOPX benchmark description

This section provides comprehensive insights regarding the ten GTOPX benchmark instances. While all issues pertaining to the original GTOP database were characterized by a singular objective and a continuous nature within the search space domain, the GTOPX compilation incorporates three novel problem instances that exhibit mixed-integer and multi-objective problem characteristics. The overarching mathematical representation of the optimization problem addressed in GTOPX is articulated as multi-objective Mixed-integer Nonlinear Programming (MINLP):

$$\text{Minimize } f_i(x, y) \ (x \in \mathbb{R}^{n_{\text{con}}}, y \in \mathbb{N}^{n_{\text{int}}}, n_{\text{con}}, n_{\text{int}} \in \mathbb{N} \tag{1}$$

Subject to: $\tag{2}$

$$O_i(x, y) \geq 0, i = 0,1,2, \dots, m \in \mathbb{N}$$
$$x_l \leq x \leq x_u (x_l, x_u \in \mathbb{R}^{n_{\text{con}}}) \text{ and}$$
$$y_l \leq y \leq y_u \ (y_l, y_u \in \mathbb{N}^{n_{\text{int}}}).$$

where $f_i(x,y)$ and $O_i(x,y)$ signify the objective and constraint functions that are contingent upon continuous (x) and discrete (y) decision variables, which are subject to box constraints delineated by specific lower ($x_l, y_l$) and upper ($x_u, y_u$) bounds, the ten discrete benchmark instances of GTOPX are enumerated in [28], inclusive of their nomenclature, the count of objectives, variables, and constraints, as well as the best-known objective function value $f(x,y)$. It is pertinent to note that due to the extensive duration for which these benchmarks have been accessible and subjected to testing, it is posited that all enumerated solutions have effectively converged [28].

## 2.1    Cassini-1

The Cassini-1 benchmark simulates an interplanetary space mission directed towards Saturn. The primary objective of this mission is to achieve capture by Saturn's gravitational field into an orbit characterized by a pericenter radius of 108,950 km and an eccentricity of 0.98. The chronological sequence of fly-by planets designated for this mission is Earth–Venus–Earth–Jupiter–Saturn, with the initial entity serving as the commencement planet and the terminal entity as the final target. The objective function associated with this benchmark is to minimize the cumulative velocity change $\Delta V$ incurred throughout the mission, encompassing both launch and capture manoeuvres. This benchmark encompasses six decision variables (refer to [28]). The most accurate solution to this benchmark is characterized by an objective function value of $f(x) = 4.9307$, and the corresponding vector of decision variable solutions $x$ is accessible online [14]. Fig.1 shows the Cassini's trajectory steps.

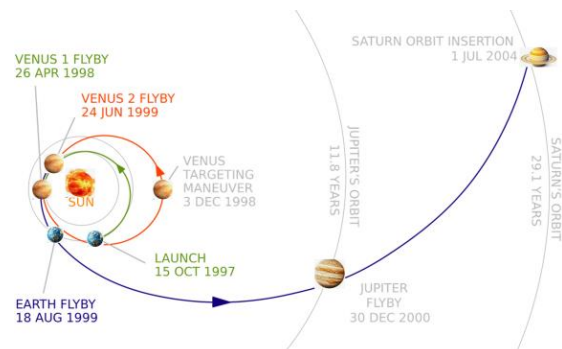

Fig.1: Cassini's trajectory is represented schematically, redesigned from [16].

## 2.2    Cassini-2

The Cassini-2 benchmark[28] models an interplanetary space mission directed towards Saturn, which includes deep space manoeuvres (DSM) and is thus significantly more complex than its predecessor benchmark, Cassini-1. The sequence of fly-by planets for this mission is also delineated as Earth–Venus–Earth–Jupiter –Saturn, where the initial item signifies the start planet and the final item represents the ultimate target. The goal of this benchmark is similarly to minimize the total $\Delta V$ accumulated throughout the entirety of the mission; however, the objective here is to facilitate a rendezvous, whereas in Cassini-1 the objective pertains to orbit insertion. This benchmark involves 22 decision variables (see [28]). The most accurate solution to this benchmark is associated with an objective function value of $f(x) = 8.3830$, and the corresponding vector of decision variable solutions $x$ is available online [14].

## 2.3    Rosetta

The Rosetta benchmark delineates multi-gravity assist space missions directed towards comet 67P/Churyumov-Gerasimenko, including deep space manoeuvres (DSM). The sequence of planetary fly-bys designated for this mission is articulated as Earth–Earth–Mars–Earth–Earth–67P. The aim of this benchmark is to minimize the total $\Delta V$ accrued throughout the mission. This benchmark encompasses 22 decision variables (refer to [28]). The most optimal solution presently known for this benchmark possesses an objective function value of $f(x) = 1.3434$, with the vector of solution decision variables x accessible online [14].

# 3    Methods

Like other algorithms, the GWO has weaknesses in obtaining optimal solutions during the search processes in some cases[24]. Thus, several investigations were presented in this study to enhance the GWO optimization processes and handle the disadvantages, such as local optima stagnation, low convergence speed, low exploration capabilities, and imbalance between exploration and exploitation. The proposed approach is predicated on the subsequent assumptions:

- All wolves' positions are updated based on the new equations that are derived from the MPA algorithm and adapted to GWO parameters.
- GMPA optimization phases are divided into three major phases in order to excel in the GWO optimization phase.
- The parameters associated with P and FADs, which remain constant within the MPA framework, undergo modifications and adaptive adjustments to the step size of the wolves' movement, thereby enabling them to effectively navigate the search space during the subsequent exploitation phase in GMPA.

## 3.1        Proposed hybrid optimisation method (GMPA)

Despite the fact that GMPA resembles the MPA optimization scenarios, the first phase of GMPA belongs to the exploration phase. The exploration phase occurs within the initial one-third of iterations and is mathematically expressed as follows:

While $t < \frac{1}{3}T$                                                                           (3)

$$\overrightarrow{\text{stepsize}} = \overrightarrow{R_B} \otimes \left( \overrightarrow{X_\alpha} - \overrightarrow{R_B} \otimes \overrightarrow{X_p}(i) \right) i = 1, \ldots, n$$

$$\overrightarrow{X_1} = \overrightarrow{X_\alpha} + P \cdot \boldsymbol{R} \otimes \overrightarrow{\text{stepsize}}_{\iota}$$

$$\overrightarrow{X_2} = \overrightarrow{X_\beta} + P \cdot \boldsymbol{R} \otimes \overrightarrow{\text{stepsize}}_{\iota}$$

$$\overrightarrow{X_3} = \overrightarrow{X_\delta} + P \cdot \boldsymbol{R} \otimes \overrightarrow{\text{stepsize}}_{\iota}$$

$$\boldsymbol{X}(i+1) = \frac{\overrightarrow{X_1} + \overrightarrow{X_2} + \overrightarrow{X_3}}{3}$$

In the course of the proposed approach's second phase, there is a gradual transition from exploration to exploitation. Consequently, the entire population is divided into two equal factions, wherein the first faction engages in exploitation while the second faction engages in exploration as follows:

While $\frac{1}{3}T < t < \frac{2}{3}T$                                                           (4)

$$\overrightarrow{\text{stepsize}}_{\iota} = \overrightarrow{R_L} \otimes \left( \overrightarrow{X_\alpha} - \overrightarrow{R_L} \otimes \overrightarrow{X_p}(i) \right) i = 1, \ldots, n/2$$

The step sizes derived from the Lévy distribution predominantly consist of minor increments, rendering them particularly advantageous for the exploitation phase. The behaviour exhibited by the second faction in GMPA is characterized as follows:

$$\overrightarrow{\text{stepsize}}_{\iota} = \overrightarrow{R_B} \otimes \left( \overrightarrow{R_B} \otimes \overrightarrow{X_\alpha} - \overrightarrow{X_p}(i) \right) i = n/2, \ldots, n$$                    (5)

$$\overrightarrow{X_p}(i) = \overrightarrow{X_\alpha} + P \cdot CF \otimes \overrightarrow{\text{stepsize}}_{\iota}, CF = \left( 1 - \frac{t}{T} \right)^{\left( 2\frac{t}{T} \right)}$$

The third phase, designated as the exploitation phase, occurs during the final one-third of iterations. This phase is mathematically delineated as follows:

While $t > \frac{2}{3}T$                                                                            (6)

$$\overrightarrow{\text{stepsize}}^2_{\iota} = \overrightarrow{R_L} \otimes \left( \overrightarrow{R_L} \otimes \overrightarrow{X_\alpha} - \overrightarrow{X_p}(i) \right) i = 1, \ldots, n$$

$$\overrightarrow{X_p}(i) = \overrightarrow{X_\alpha} + P \cdot CF \otimes \overrightarrow{\text{stepsize}}_{\iota}$$

The incorporation of the step size into the alpha position signifies the modelling of the wolves' movement, aimed at facilitating the updating of the prey's position. FADs, along with long skips, serve to avert stagnation of the MPA within local optima, thereby enhancing the efficacy of the algorithm. Nevertheless, in instances where the MPA becomes ensnared in a local optimal, the long skips may compel the MPA to invest a significant portion of its iterations in that region, precluding the opportunity to explore the solution space for superior solutions and ultimately achieving a global optimal. Thus, the FAD scenario within GMPA can be articulated as follows:

$$\overrightarrow{X_p}(i) = \begin{cases} \overrightarrow{X_p}(i) + CF[\boldsymbol{X}_{\min} + \boldsymbol{R} \otimes (\boldsymbol{X}_{\max} - \boldsymbol{X}_{\min})] \otimes \boldsymbol{U} & \text{if } r1 \leq r2 \\ \overrightarrow{X_p}(i) + [FADS(1-r) + r](\overrightarrow{Xp_{r1}} - \overrightarrow{Xp_{r2}}) & \text{if } r1 > r2 \end{cases} \qquad (7)$$

The GMPA possesses the capability to retain the optimal solution acquired in a prior iteration and juxtapose it with the current solution. By harnessing this capacity, the GMPA is enabled to systematically diminish the inferior solutions throughout the optimization phase and to investigate potential regions within the solution space where the likelihood of locating the global optimum is markedly elevated.

Generate the local alpha wolf position matric in the proposed method, in order to better search the solution space of the problem, a neighbourhood matrix based on the position of the alpha wolf (the best solution) is generated at the end of each iteration. The purpose of generating this matrix is twofold: first, to perform a quick local search to find the best possible solution, and second, to update the fitness matrix as well as the positions of the alpha, beta, and delta wolves at the end of each iteration, which leads to a better search for the solution space to find the global optimal solution. This step is formulated in the proposed algorithm as follows. The utilization of diminutive values $\tau$ facilitates exploration within a constrained range, whereas substantial values for this parameter permit investigation across an expansive range. Indeed, conducting searches over a broad spectrum yields a heightened diversity of solutions and augments the likelihood of identifying near-optimal solutions (global search). As the algorithm's outputs converge towards the global optimal solution, engaging in a focused search within a limited range enhances the precision of locating the global optimal solution (local search), thereby underscoring the necessity of achieving a harmonious equilibrium between these two functionalities. The rationale for this is that reliance solely on global search operators results in the algorithm's inability to ascertain the global optimum with the requisite precision. Conversely, exclusive reliance on local search renders the algorithm susceptible to entrapment within local optima. Consequently, to establish a judicious equilibrium between global and local search methodologies, the current investigation employs the following formula by systematically decreasing $\tau$ from a substantial value to a minimal value. Thus, the direction of $\tau$ is oriented towards a randomized position to foster greater diversity.

$$\tau = W * \left( \text{rand} * \varphi - \text{rand} * \overrightarrow{X_p}(i) \right) * \text{norm}\left( \overrightarrow{X_\alpha} - \overrightarrow{X_p}(i) \right) \qquad (8)$$

where rand denotes a stochastic variable with a uniform distribution, $\varphi$ refers to a randomly generated position as delineated by follows.

$$\varphi = lb + \text{rand}(1,d) * (ub - lb) \tag{9}$$

$W$ signifies a nonlinear weight represented by a stochastic variable ranging between 0 and infinity. In this context, the initial term indicates that $\overrightarrow{X_\alpha}$ transitions towards a random position ($\varphi$). Regarding the subsequent term, as iterations progress, $\overrightarrow{X_p}(i)$ approaches $\overrightarrow{X_\alpha}$, resulting in the Euclidean distance between $\overrightarrow{X_p}(i)$ and $\overrightarrow{X_\alpha}$ diminishing towards zero. Consequently, the local search is initiated. The computation of $W$ in the third term is articulated as follows:

$$W = \left( \left( \left( 1 - 1 * \frac{t}{T} + \varepsilon \right)^{2*\text{randn}} \right) * \left( \text{rand}(1,d) * \frac{t}{T} \right) * \text{rand}(1,d) \right) \tag{10}$$

From Eq. 10, it is evident that W exhibits significant variability throughout the iterative process. This characteristic ensures the evasion from local optima within the proposed methodology. Finally, the alpha's neighbor's matric is generated as follows:

$$\overrightarrow{NX_\alpha} = \overrightarrow{X_\alpha} + \text{randn}(1,d) * \tau \tag{11}$$

where $NX_\alpha$ signifies the neighbouring entity, and randn represents a stochastic variable derived from a normal distribution characterized by a mean of zero and a standard deviation of one. The pseudo-code of GMPA is delineated in Algorithm 3.

## 4    Experimental Results

In this scholarly investigation, we provide an exhaustive comparative analysis of the recently introduced hybrid optimization methodology, GMPA, juxtaposed against a spectrum of well-established algorithms—specifically MPA [4], GWO [8], DE, LSHADE [7], PSO [6] —utilizing the CEC 2022 benchmark suite as our evaluation framework. The CEC 2022 suite comprises an array of intricate and formidable benchmark functions meticulously crafted to evaluate the efficacy of optimization algorithms operating within high-dimensional, multimodal, and stochastic environments [12]. The benchmark encompasses diverse functions that rigorously assess an algorithm's capacity to discern global optima while circumventing local minima. These functions encompass both continuous and multi-objective challenges characterized by varying levels of complexity, such as substantial quantities of local optima, non-linearity, and high dimensionality. The CEC 2022 benchmark is extensively recognized for its utility in the assessment and comparative analysis of optimization methodologies, rendering it an exemplary platform for evaluating the performance of GMPA.

Statistical significance assessments further substantiate the preeminence of GMPA. The p-values, derived from a comparative analysis of the results obtained by GMPA and the other optimization methodologies, consistently denote that the performance of GMPA is statistically significant. For the majority of functions, the p-values remain well below the 0.05 threshold, thereby affirming that the observed performance discrepancies are not attributable to random variation. For example, in functions F1, F2, F3, and F6, the p-values for GMPA are exceedingly low (ranging from 3.02E-11 to 2.92E09), thereby accentuating the method's superior performance in relation to MPA, GWO, DE, and LSHADE. These findings suggest that GMPA not only demonstrates greater efficacy in identifying optimal solutions but also manifests enhanced robustness and consistency when confronted with challenging optimization scenarios. In contrast, PSO, GWO, DE, and LSHADE display higher $p$-values in these functions, signifying that their performance lags in comparison to GMPA in terms of both solution quality and stability.

Further examination of the efficacy of GMPA across all benchmark functions elucidates its resilience in both unimodal and multimodal contexts. In functions such as F2, F4, and F5, GMPA exhibits lower AVG and STD values in comparison to MPA, GWO, and DE. For instance, in F2, GMPA attains an AVG of $4.15E + 02$, markedly surpassing DE (4.49E+02) and GWO (5.02E+02). In more complex functions like F6, which is significantly multimodal, GMPA persists in its superior performance, achieving an AVG of $1.81E + 03$, in contrast to DE(6.24E + 05) and LSHADE(5.68E + 05). This performance is emblematic of GMPA's proficient exploration of the search space and its ability to circumvent local optima, rendering it a particularly robust methodology for intricate, high-dimensional optimization challenges. Moreover, in F12, a function characterized by noise and multimodality, GMPA demonstrates a distinct advantage over competing methods, achieving an AVG of $2.90E + 03$ with a reduced STD ( $5.36E - 05$ ), while alternative methodologies such as PSO and DE reveal larger STD values, indicative of greater variability and diminished consistency in performance.

Regarding statistical performance that is reported in Table 1, GMPA consistently surpasses the other algorithms across the majority of the evaluated functions, including MPA, GWO, DE, and LSHADE, among others. For instance, in the CEC 2022 F1 function, which is classified as unimodal, GMPA attains the most favourable average (AVG) value of $3.00E + 02$, eclipsing the performances of GWO (1.09E+04) and DE(1.53E+04), among others. Furthermore, the standard deviation (STD) associated with GMPA is the lowest, thereby indicating its enhanced consistency and robustness in successfully identifying the global optimum across multiple iterations. Likewise, in the more intricate F6 and F12 functions, characterized by high multimodality, GMPA sustains its advantage in terms of AVG, achieving values of $1.81E+03$ and $2.90E+03$, respectively, whereas methodologies such as DE and LSHADE exhibit significantly elevated AVG values, suggesting a less effective exploration of the search space in F6. In the case of F6, the standard deviation of GMPA ( $2.98E+01$ ) is markedly smaller than that of LSHADE $(1.31E + 06)$ and DE(3.10E + 05), thereby underscoring the method's superior stability and reliability in the face of challenging conditions. In conclusion, the GMPA methodology proves to be exceptionally effective when evaluated against the CEC 2022 benchmark suite, showcasing superior performance for Table 1: Statistical results of 12 CEC optimisation benchmarks using the hybrid proposed method (GMPA) compared with five other methods.

Table 1: Statistical results of 12 CEC optimisation benchmarks using the hybrid proposed method (GMPA) compared with five other methods.

| Methods: | | GMPA | GWO | PSO | MPA | DE | LSHADE | Methods: | | GMPA | GWO | PSO | MPA | DE | LSHADE |
|---|---|---|---|---|---|---|---|---|---|---|---|---|---|---|---|
| F1 | AVG | 3.00E+02 | 1.09E+04 | 3.00E+02 | 3.00E+02 | 1.53E+04 | 3.54E+04 | F2 | AVG | 4.15E+02 | 5.02E+02 | 4.26E+02 | 4.49E+02 | 4.49E+02 | 4.17E+02 |
| | STD | 6.74E-02 | 4.66E+03 | 4.27E-05 | 1.96E-03 | 3.44E+03 | 3.01E+04 | | STD | 1.69E+01 | 5.36E+01 | 2.39E-01 | 1.49E+01 | 7.58E-01 | 1.51E+00 |
| | Min | 3.00E+02 | 2.33E+03 | 3.00E+02 | 3.00E+02 | 9.93E+03 | 3.42E+02 | | Min | 4.00E+02 | 4.00E+02 | 4.08E+02 | 4.03E+02 | 4.45E+02 | 4.13E+02 |
| | Max | 3.00E+02 | 2.17E+04 | 3.00E+02 | 3.00E+02 | 2.35E+04 | 1.01E+05 | | Max | 4.49E+02 | 6.51E+02 | 4.75E+02 | 4.71E+02 | 4.49E+02 | 4.29E+02 |
| | P-Value | | 3.02E-11 | 3.02E-11 | 1.21E-10 | 3.02E-11 | 3.02E-11 | | P-Value | | 3.02E-11 | 4.21E-02 | 3.92E-02 | 7.09E-08 | 6.77E-05 |
| F3 | AVG | 6.00E+02 | 6.05E+02 | 6.35E+02 | 6.00E+02 | 6.00E+02 | 6.00E+02 | F4 | AVG | 8.08E+02 | 8.52E+02 | 8.63E+02 | 8.36E+02 | 8.93E+02 | 8.56E+02 |
| | STD | 1.92E-03 | 3.96E+00 | 1.03E+01 | 4.92E-02 | 0.00E+00 | 6.17E-02 | | STD | 1.37E+01 | 2.37E+01 | 1.74E+01 | 7.92E+00 | 1.08E+01 | 3.53E+01 |
| | Min | 6.00E+02 | 6.01E+02 | 6.16E+02 | 6.00E+02 | 6.00E+02 | 6.00E+02 | | Min | 8.02E+02 | 8.22E+02 | 8.43E+02 | 8.17E+02 | 8.67E+02 | 8.19E+02 |
| | Max | 6.00E+02 | 6.17E+02 | 6.54E+02 | 6.00E+02 | 6.00E+02 | 6.00E+02 | | Max | 8.66E+02 | 9.30E+02 | 9.09E+02 | 8.53E+02 | 9.13E+02 | 9.28E+02 |
| | P-Value | | 3.02E-11 | 3.02E-11 | 1.46E-10 | 1.21E-12 | 1.86E-03 | | P-Value | | 1.33E-02 | 3.08E-08 | 7.62E-01 | 3.02E-11 | 1.02E-01 |
| F5 | AVG | 9.01E+02 | 1.16E-03 | 1.74E+03 | 9.25E+02 | 9.00E+02 | 9.06E+02 | F6 | AVG | 1.81E+03 | 1.10E-06 | 3.52E-03 | 1.82E+03 | 6.24E-05 | 6.68E-05 |
| | STD | 2.38E+01 | 2.86E+02 | 1.39E+02 | 2.36E+01 | 4.03E-01 | 2.08E+01 | | STD | 2.84E+03 | 1.54E+03 | 5.66E+06 | 2.07E+03 | 9.12E+03 | 1.31E+06 |
| | Min | 9.03E+02 | 9.32E+02 | 9.01E+02 | 9.00E+02 | 9.00E+02 | 9.00E+02 | | Min | 1.80E+02 | 2.22E+03 | 1.91E+03 | 1.80E+03 | 6.44E+04 | 2.33E+03 |
| | Max | 9.93E+02 | 6.17E+02 | 2.57E+03 | 1.03E+03 | 9.02E+02 | 9.97E+02 | | Max | 1.92E+03 | 2.78E+03 | 7.57E+03 | 1.84E+03 | 1.32E+06 | 4.97E+06 |
| | P-Value | | 2.92E-09 | 8.48E-09 | 1.67E-01 | 3.02E-11 | 1.31E-08 | | P-Value | | 1.55E-09 | 2.83E-08 | 7.22E-06 | 3.47E-10 |
| F7 | AVG | 2.01E+03 | 2.08E+03 | 2.13E+03 | 2.03E+03 | 2.03E+03 | 2.04E+03 | F8 | AVG | 2.22E+03 | 2.26E+03 | 2.31E+03 | 2.22E+03 | 2.23E+03 | 2.23E+03 |
| | STD | 1.06E+01 | 4.04E+01 | 5.35E+01 | 5.35E+00 | 4.22E+00 | 1.70E+01 | | STD | 2.23E+00 | 3.35E+01 | 6.91E+02 | 2.81E+00 | 1.01E+01 | 1.62E+00 |
| | Min | 2.00E+03 | 2.02E+03 | 2.06E+03 | 2.02E+03 | 2.03E+03 | 2.02E+03 | | Min | 2.22E+03 | 2.22E+03 | 2.26E+03 | 2.22E+03 | 2.22E+03 | 2.22E+03 |
| | Max | 2.06E+03 | 2.21E+03 | 2.24E+03 | 2.05E+03 | 2.04E+03 | 2.10E+03 | | Max | 2.22E+03 | 2.35E+03 | 2.56E+03 | 2.22E+03 | 2.23E+03 | 2.24E+03 |
| | P-Value | | 2.02E-08 | 1.78E-10 | 7.39E-01 | 4.06E-02 | 1.19E-01 | | P-Value | | 7.69E-08 | 4.83E-01 | 3.27E-03 | 3.02E-11 |
| F9 | AVG | 2.45E+03 | 2.51E+03 | 2.47E+03 | 2.48E+03 | 2.48E+03 | 2.47E+03 | F10 | AVG | 2.50E+03 | 3.30E+03 | 3.97E+03 | 2.50E+03 | 2.48E+03 | 3.14E+03 |
| | STD | 7.62E-05 | 2.46E+01 | 1.33E-07 | 9.92E-07 | 7.68E-11 | 1.02E+01 | | STD | 9.60E+00 | 6.91E+02 | 8.41E+02 | 1.13E+00 | 2.98E+01 | 9.84E+02 |
| | Min | 2.42E+03 | 2.48E+03 | 2.47E+03 | 2.48E+03 | 2.48E+03 | 2.47E+03 | | Min | 2.45E+03 | 2.50E+03 | 2.50E+03 | 2.50E+03 | 2.41E+03 | 2.40E+03 |
| | Max | 2.48E+03 | 2.58E+03 | 2.47E+03 | 2.48E+03 | 2.48E+03 | 2.47E+03 | | Max | 2.93E+03 | 6.49E+03 | 4.30E+03 | 2.50E+03 | 2.51E+03 | 5.67E+03 |
| | P-Value | | 3.02E-11 | 1.02E-11 | 9.82E-01 | 3.02E-11 | 3.02E-11 | | P-Value | | 5.60E-07 | 7.09E-08 | 2.84E-01 | 4.12E-01 |
| F11 | AVG | 2.92E+03 | 3.43E+03 | 2.95E+03 | 2.90E+03 | 2.90E+03 | 2.90E+03 | F12 | AVG | 2.90E+03 | 2.97E+03 | 3.23E+03 | 2.94E+03 | 2.90E+03 | 2.90E+03 |
| | STD | 4.04E+01 | 1.10E+02 | 1.56E+02 | 9.69E-01 | 1.06E+02 | | | STD | 5.36E-05 | 2.31E+01 | 2.60E+02 | 3.70E+00 | 2.32E+00 | 2.16E-04 |
| | Min | 2.90E+03 | 3.13E+03 | 2.60E+03 | 2.60E+03 | 2.90E+03 | 2.90E+03 | | Min | 2.90E+03 | 2.95E+03 | 2.89E+03 | 2.93E+03 | 2.90E+03 | 2.90E+03 |
| | Max | 2.90E+03 | 3.83E+03 | 3.35E+03 | 3.06E+03 | 3.90E+03 | 3.32E+03 | | Max | 2.90E+03 | 3.05E+03 | 3.62E+03 | 2.95E+03 | 2.90E+03 | 2.90E+03 |
| | P-Value | | 3.02E-11 | 1.29E-02 | 1.33E-01 | 1.17E-09 | 3.15E-02 | | P-Value | | 4.57E-09 | 7.96E-03 | 4.71E-04 | 2.00E-05 | 3.02E-11 |

In terms of both solution quality and stability in relation to MPA, GWO, DE, LSHADE, and other leading optimization algorithms. The amalgamation of low AVG, low STD, and statistically significant p-values across a diverse array of challenging benchmark functions positions GMPA as a formidable candidate for addressing complex optimization issues in practical applications. These findings underscore the significance of hybridizing diverse optimization strategies, as exemplified by the GMPA method, which capitalizes on the strengths of both the GWO and the MPA to achieve enhanced performance.

Fig 2 shows the convergence analysis further substantiates the remarkable performance of GMPA, showcasing its capacity to achieve rapid convergence to high-quality solutions across a range of objective functions, specifically CF1, CF2, and CF6. In comparison to alternative methods, GMPA consistently realizes accelerated convergence rates along with superior final solutions, thereby emphasizing its robust adaptability to a variety of optimization landscapes. For example, in the cases of CF1 and CF2, GMPA stabilizes within the initial 500 iterations, while algorithms such as GWO and LSHADE experience significantly slower convergence, frequently stalling at suboptimal solutions. Even when addressing more intricate functions like CF11, GMPA leads with enhanced convergence quality, whereas other methodologies, such as GWP, fall behind, demonstrating slower advancement and diminished exploitation capabilities. These results accentuate the potential of hybrid meta-heuristic strategies, exemplified by GMPA, in tackling complex optimization challenges, presenting further opportunities for refinement and implementation within real-world problem contexts. Finally, in the context of Function F7, which presents a highly sensitive optimization landscape, GMPA attains an average of 1.03E+07, significantly outperforming methods such as PSO, DE, and LSHADE, which yield suboptimal performance with average values approaching. The reliability of GMPA is further substantiated by its low standard deviation of 1.49E+07, rendering it more stable and dependable in such demanding environments. Additionally, the p-value affirms the statistical significance of these findings. In summary, the thorough assessment of GMPA within the GTOPX benchmark suite elucidates its considerable advantages in comparison to traditional optimization methodologies, including GWO, MPA, PSO, DE, and LSHADE. The amalgamation of GWO and MPA engenders a formidable synergy, ensuring rapid convergence, enhanced constraint management, and augmented robustness, particularly in the context of high-dimensional, multimodal, noisy, and mixed-integer optimization problems. The consistently low average values, minimal standard deviations, and statistically significant p-values from the results accentuate the superior performance of GMPA, establishing it as a commendable option for addressing intricate optimization challenges across a diverse array of applications. In

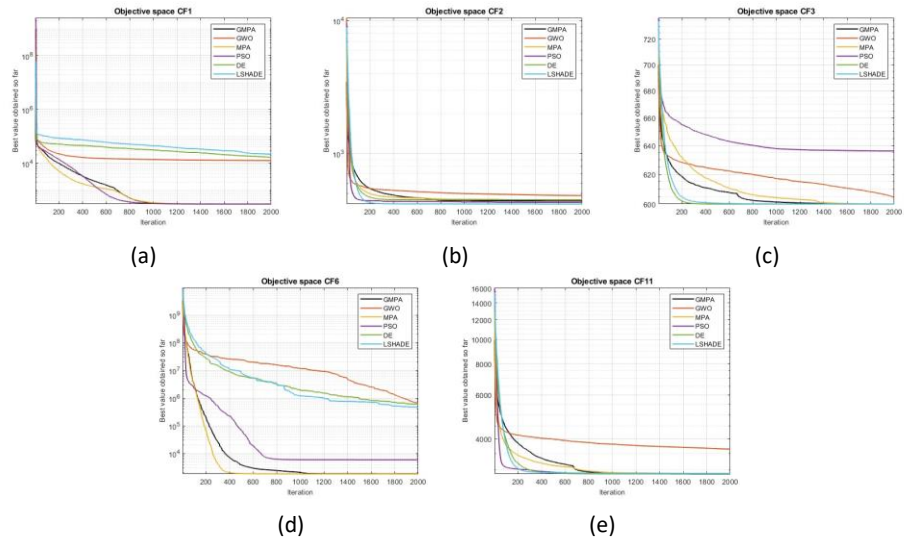

Fig.2: A convergence comparison for five CEC benchmarks and the proposed method compared with the other five methods.

the present research endeavour, the efficacy of the proposed hybrid optimization algorithm, GMPA, is meticulously assessed using the GTOPX benchmark suite and is juxtaposed with various cutting-edge optimization methodologies, including GWO,

MPA, DE, PSO and LSHADE. The GTOPX benchmark is extensively acknowledged as an exemplary test suite for the evaluation of optimization algorithms, as it encompasses a heterogeneous array of interplanetary space trajectory design challenges, ranging from unimodal to intricately multimodal problems, characterized by diverse levels of non-linearity, mixed-integer variables, and multiple objectives. These challenges rigorously test the capacity of optimization algorithms to effectively navigate the dual demands of exploration and exploitation within high-dimensional, noisy, and constrained search spaces, thereby establishing GTOPX as a formidable platform for the benchmarking of GMPA. The outcomes derived from the GTOPX benchmark in Table 3 unequivocally illustrate the superior performance of GMPA across an extensive majority of the test cases, accentuating its proficiency in attaining optimal solutions while exhibiting stability across multiple iterations. For instance, in the Cassini1 problem (a comparatively straightforward unimodal challenge), GMPA realizes the lowest average (AVG) value of 9.62, thus surpassing GWO (12.5) and PSO (19.7). Furthermore, it also sustains a reduced standard deviation, indicative of enhanced consistency throughout the search endeavour. The statistical significance of these findings is corroborated by the p-value, implying that the observed performance enhancement is attributable to substantive factors rather than mere chance. Moreover, when evaluating high-dimensional noisy functions such as F5 and F6, GMPA continues to exhibit superior performance. In the case of F5, characterized by substantial negative values and intricate dynamics, GMPA achieves an average of -7.36E+05, outperforming alternative methodologies such as DE (-6.31E+05) and MPA (-9.96E+05) while demonstrating a reduced standard deviation of 3.55E-10. Similarly, for F6, a more intricate multimodal challenge, GMPA attains an average of 8.70, eclipsing the performances of GWO (13.9) and PSO (9.92) while maintaining an almost negligible standard deviation of 1.81E-15, thereby underscoring its stability and resilience across various test instances. Conversely, methodologies such as GWO and PSO exhibit difficulties characterized by heightened variability and premature convergence when confronted with these noisy, high-dimensional challenges. The hybridization of GMPA unequivocally demonstrates its efficacy in balancing the exploration of novel regions with the exploitation of established areas, thus enabling it to circumvent local optima more effectively than conventional approaches. Fig 3 delineates the convergence analysis of the GMPA algorithm in comparison to five other optimization methodologies—namely GWO, MPA, PSO, DE, and LSHADE—across three benchmark functions (Cassini 1, Cassini 2, and Rosetta) derived from the GTOPX dataset. The convergence trajectories, illustrated on a logarithmic scale, indicate that GMPA consistently surpasses the competing techniques in terms of both convergence velocity and the quality of the final solutions. For Cassini 1, GMPA realizes swift convergence within the initial few hundred iterations and achieves the optimal objective values, thereby emphasizing its superior balance between exploration and exploitation. In the cases of Cassini 2, GMPA exhibits a continuous decrease in the objective function values, significantly outpacing the performance of MPA and DE, which exhibit early stagnation. Likewise, for Rosetta, GMPA showcases the most proficient convergence, attaining the lowest error values among all evaluated methods. These findings underscore GMPA's robustness and adaptability in tackling a variety of optimization challenges, thereby affirming its efficacy across diverse problem landscapes. The empirical evaluation of six meta-heuristic algorithms, comprising GMPA, GWO, MPA, PSO, DE, and LSHADE, conducted on the GTOPX benchmark dataset, elucidates profound insights regarding their optimization efficacy.

Across a spectrum of seven benchmark problems that is pictured in Fig. 4, GMPA virtually surpasses alternative methodologies, attaining the lowest median values and exhibiting a narrow interquartile range (IQR), which signifies a high degree of reliability and consistent performance.

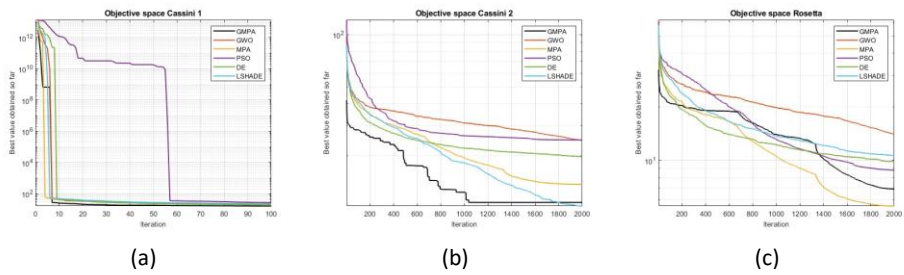

Fig.3: GMPA versus other approaches, convergence analysis on GTOPX

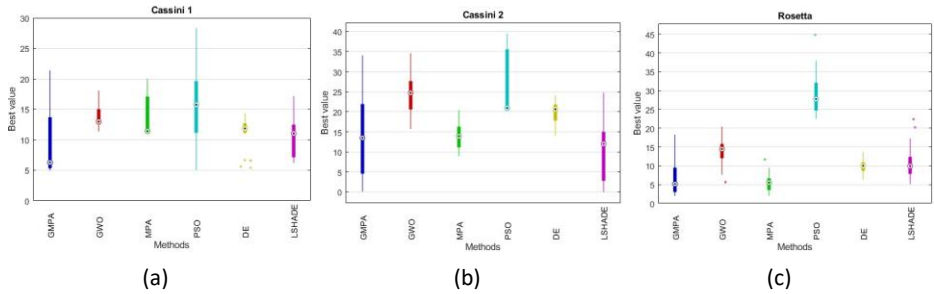

Fig.4: GMPA versus other approaches, convergence analysis on GTOPX

Remarkably, in the context of challenging benchmarks, precisely Benchmark 1 and Benchmark 6, GMPA displays superior consistency, while algorithms such as PSO and MPA also reveal competitive yet comparatively less reliable outcomes. In contrast, GWO manifests considerable variability, as illustrated by extended whiskers and recurrent outliers, highlighting its vulnerability to inconsistent optimization results.

Benchmarks characterized by increased complexity, such as Benchmark 4, demonstrate heightened variability across all methodologies; nonetheless, GMPA persists in maintaining its competitive advantage, thereby underscoring its effective balance between exploration and exploitation.

## 5    Conclusion

In this research, a novel and efficacious mechanism designated as GMPA has been introduced, which facilitates the identification of optimal trajectories for space missions intended for the transmission of data generated by applications within the domain of FinTech related to outer space endeavours. Initially, the manuscript undertakes a comprehensive examination of the limitations inherent in contemporary methodologies concerning single-objective-constrained optimization challenges as evaluated against the GTOPX and CEC 2022 benchmarks. Our analysis has revealed that the deployment of sufficiently sophisticated optimization algorithms is paramount in effectively and efficiently deriving optimal solutions to such intricate problems. Consequently, we have devised a hybrid GWO MPA optimization algorithm that enhances the performance and quality of the solutions selected for the specified challenges. This newly proposed algorithm was initially subjected to rigorous testing against a diverse array of benchmark functions (CEC 2022) to assess and substantiate its capacity to attain the global optimal solution amidst a collection of available local optimal solutions. Subsequently, the efficacy of GMPA was evaluated in the context of the GTOPX challenges, which epitomize genuine interplanetary trajectory design issues. Our proposed optimization algorithm has been executed in conjunction with other prominent algorithms to address seven distinct problems within the GTOPX benchmark. The results indicate that our algorithm consistently outperformed its counterparts in numerous instances when identifying the optimal solution.